\PassOptionsToPackage{dvipsnames}{xcolor}
\documentclass[11pt]{article}

\usepackage[preprint]{acl}

\usepackage{times}
\usepackage{latexsym}

\usepackage[T1]{fontenc}

\usepackage[utf8]{inputenc}

\usepackage{microtype}
\usepackage{booktabs}
\usepackage{graphicx}
\usepackage{amsmath}
\usepackage{amssymb}
\usepackage{algorithm}
\usepackage{algpseudocode}
\usepackage{enumitem}
\usepackage{listings}
\usepackage{needspace}
\usepackage{tcolorbox}
\tcbuselibrary{listings,breakable,skins}
\usepackage{tabularx}
\usepackage{subcaption}
\usepackage{xcolor}

\newtcblisting{promptbox}[2][]{   
    breakable,
    enhanced,
    colback=gray!5,
    colframe=Orchid!60,
    colbacktitle=Orchid!40,
    coltitle=black,
    fonttitle=\bfseries\small,
    title={#2},
    title after break={#2 (cont.)},   
    boxrule=0.4pt,
    arc=1.5pt,
    left=4pt, right=4pt, top=3pt, bottom=3pt,
    toptitle=1.5pt, bottomtitle=1.5pt,
    listing only,
    listing options={
        breaklines=true,
        breakatwhitespace=true,
        columns=fullflexible,
        keepspaces=true,
        basicstyle={\scriptsize\ttfamily\linespread{0.9}\selectfont},
        xleftmargin=0pt,
        xrightmargin=0pt
    },
    #1   
}

\usepackage{amsmath,amsfonts,bm}

\def\eqref#1{equation~\ref{#1}}

\def\1{\bm{1}}

\DeclareMathAlphabet{\mathsfit}{\encodingdefault}{\sfdefault}{m}{sl}
\SetMathAlphabet{\mathsfit}{bold}{\encodingdefault}{\sfdefault}{bx}{n}

\lstdefinelanguage{Markdown}{
    basicstyle=\ttfamily\footnotesize,
}

\title{Revision-Based Learning from Interactive Experience and Feedback}

\author{
  \textbf{Jiawen Li}$^{1}$ \quad
  \textbf{Chenming Ge}$^{2}$ \quad
  \textbf{Alice Gao}$^{3}$ \\
  \textbf{Rishab Vempati}$^{1}$ \quad
  \textbf{Yi-Hao Peng}$^{4}$ \quad
  \textbf{Jason Wu}$^{1}$ \\[0.5em]
  $^{1}$Purdue University \quad
  $^{2}$University of Michigan \\
  $^{3}$University of Washington \quad
  $^{4}$MIT \\[0.5em]
  \texttt{\{jiawenli,rvempat,jasonwu\}@purdue.edu} \\
  \texttt{gecm@umich.edu} \quad
  \texttt{atgao@cs.washington.edu} \quad
  \texttt{yihao@csail.mit.edu}
}

\begin{document}

\maketitle

\begin{abstract}

Code generation models are increasingly used to produce interactive artifacts, such as websites and user interfaces. Many properties that determine whether these interfaces are useful only become apparent through realistic interaction, making them difficult to incorporate into model training. In this paper, we introduce \textbf{RELIEF} (\textbf{RE}vision-based \textbf{L}earning from \textbf{I}nteractive \textbf{E}xperience and \textbf{F}eedback), a method for improving code generation models using feedback on their generated interfaces. RELIEF uses a coding agent to revise these interfaces, then incorporates the improvements into their original generation traces to produce supervision for fine-tuning. To evaluate RELIEF, we collected a dataset of 200 synthetically generated websites paired with 355 interaction recordings and accompanying feedback from 24 users. We used critiques derived from code inspection, automated interaction, explicit human feedback, and human interaction recordings to fine-tune variants of two code generation models. Our results show that feedback effectiveness varies with base model capabilities, with critiques inferred from human interaction recordings leading to the largest usability improvements overall under our tested models.

\end{abstract}

\section{Introduction}
\label{sec:introduction}

\begin{figure}[!htb]
    \centering
    \includegraphics[
        width=\columnwidth,
        height=0.27\textheight,
        keepaspectratio
    ]{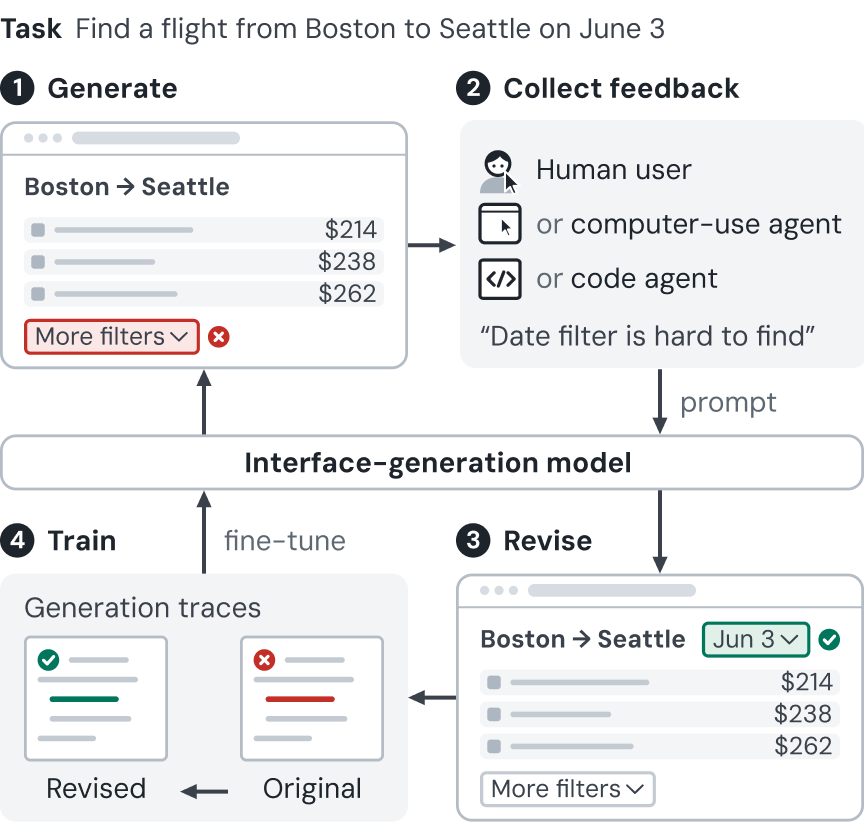}
    \caption{\textsc{RELIEF} revises generated interfaces using feedback from human users, computer-use agents, or code agents. It then reconstructs generation traces that would have produced the revised interfaces and fine-tunes the interface-generation model on these traces.}
    \label{fig:overview}
\end{figure}

Code generation models are increasingly used to produce interactive artifacts, such as websites and user interfaces, both in software development \cite{openai2025codex,autocoder2026} and dynamically within end-user applications \cite{openai2026intelligentui,chen2026generativeinterfaces}. The quality of these artifacts depends not only on functional correctness but also on how well they support people's tasks.


While properties such as syntactic validity and functional correctness can be checked automatically~\cite{chen2021codex,jimenez2024swebench}, usability is harder to evaluate. Critiques, user feedback~\cite{bennett1979commercial}, and recordings of task interactions can help identify usability problems, but these findings are typically used to improve individual interfaces, as in usability testing~\cite{dumas1993practical} and iterative test-time refinement ~\cite{madaan2023selfrefine}. Incorporating them into model training is difficult, since feedback may not translate into a reward signal, and improving an artifact alone does not directly provide an example of how the model should have generated it.

In this paper, we introduce \textbf{RELIEF} (\textbf{RE}vision-based \textbf{L}earning from \textbf{I}nteractive \textbf{E}xperience and \textbf{F}eedback), a method for generating training data from feedback on interfaces produced by coding agents. RELIEF uses a coding agent to revise the interface based on feedback, then incorporates the changes into the original generation trace. The resulting trace captures how the improved interface could have been generated from the initial prompt and serves as supervision for SFT.

To demonstrate how \textsc{RELIEF} could learn from data collected as people use model-generated interfaces, we collected 355 interaction recordings and accompanying feedback from 24 users across 200 synthetically generated websites. We applied \textsc{RELIEF} to two code generation models, Qwen 27B \cite{qwen2026qwen36} and Inkling \cite{thinkingmachines2026inkling}, using critiques derived from code inspection, automated interaction, explicit user feedback, and automated labelings of human interaction recordings. We fine-tuned the base models using these feedback sources and evaluated them on held-out generation tasks, measuring static validity through compilation and lint checks, functional correctness through computer-use-agent task completion, and usability through human preferences in a model evaluation arena. Our results show that the benefits of different feedback sources vary across base models. Among the sources evaluated, critiques inferred from human interaction recordings achieved the highest observed Elo ratings in both model families.

In summary, we contribute what is, to our knowledge, the first dataset of model-generated interfaces paired with real human interaction recordings and usability feedback, with 200 websites and 355 recordings from 24 users. We introduce \textsc{RELIEF}, which turns feedback-guided interface revisions into training supervision through generation trace reconstruction. Experiments with two open-weight code-generation models show that feedback inferred from human interaction recordings achieved the highest observed arena Elo ratings among the four feedback sources for both models. We plan to publicly release our dataset and fine-tuned model weights to support reproducibility and further research.

\section{Related Work}
\label{sec:related-work}


\paragraph{Generative UI Models.}
Recent advances in LLMs have enabled automated generation of websites and interfaces, from design-to-code translation~\citep{si2025design2code} and complete interactive website generation~\citep{lu2025webgen} to end-user applications~\citep{peng2025morae,chen2026generativeinterfaces,peng2026efficient}.
Improving these models has motivated training methods that incorporate feedback about UI quality.
Prior work has used compiler and LLM feedback~\citep{wu-2024-uicoder,li2026relook} and computer-use-agent assessments~\citep{lu2025webgenagent} to improve UI generation.
Designer comments, sketches, and revisions have also been used to train reward models and fine-tune UI generators~\citep{wu2026designer}.


\paragraph{Interface Evaluation.}
The usefulness of such feedback depends in part on how well evaluations capture the quality of completed interfaces.
Traditional usability testing and expert evaluation identify interface problems~\citep{gould1985usability,nielsen1990heuristic}, while other approaches, such as WebDev Arena~\citep{vichare2025webdevarena} and Design Arena~\citep{designarena2026}, collect preferences from end users in an ``arena'' evaluation format.
Automated evaluation has also been explored~\citep{ivory2001state,paganelli2003tools}, with recent approaches using LLM-based judgments~\citep{liu2026webcoderbench,li2026webdevjudge}, interface-specific encoders~\cite{wu2024uiclip,peng2025designpref}, computer-use agents~\citep{lin2025computer,gao2026training}, and simulated users~\citep{lu2025uxagent,xiang2024chi}.
However, predicting how interface designs affect actual user behavior remains challenging~\citep{jeon-etal-2026-mllms}.




\paragraph{Learning from Interactions.}
Feedback that arises from natural model use offers another potential source of training supervision.
Early work explored improving dialogue models using ratings and free-form feedback collected during deployment~\citep{xu-etal-2023-learning}.
ReSpect~\citep{chen2025respect} infers implicit feedback from subsequent user interactions, while WildFeedback~\citep{shi2026wildfeedback} constructs preference data from in-situ conversational feedback.
WildReward~\citep{peng-etal-2026-wildreward} further demonstrates how interaction data can be used to train reward models.
Other approaches learn from feedback-guided corrections: Juicer~\citep{shi-etal-2024-life} constructs corrected dialogue responses, TriPosT~\citep{yu-etal-2024-teaching} learns from revised reasoning trajectories, and WebWorld~\citep{wu2026webworld} uses browser-verified code revisions.
RELIEF uses feedback on completed interfaces, including observations of their use, to guide repairs that are then incorporated into the coding agents' original generation traces.

\section{Dataset}
\label{sec:dataset}


To study how feedback on generated interfaces can improve the underlying code generation models, we constructed a dataset of websites and usability critiques. In this section, we describe how we i) generated websites using two models we aimed to improve and ii) collect feedback from human users and automated agents on the model-generated outputs.

\begin{figure*}[!htb]
    \centering
    \includegraphics[width=\textwidth]{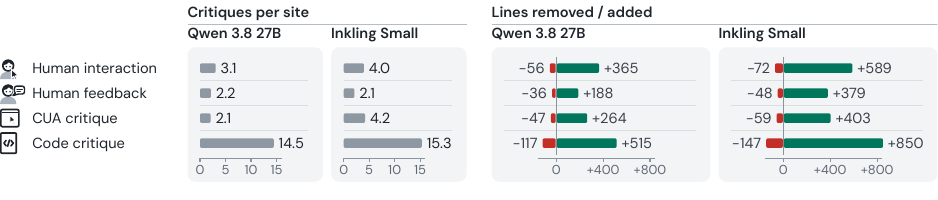}
    \caption{Feedback and code revisions per website, grouped by base model. We measure feedback volume by the number of written comments for human feedback and identified problems for other sources, and revision size by lines of code added and removed. Code agents identify several times more problems than other sources and lead to the largest revisions. Human interaction recordings also yield more critiques and larger revisions than written human feedback.}
    \label{fig:data_stats}
\end{figure*}

\subsection{Interface Dataset Generation}
We first generated a dataset of interactive websites to collect
feedback on the outputs of the code generation models we sought
to improve. We adapted the synthetic website generation procedure
of \citet{gao2026training}, drawing website URLs and task
descriptions from Mind2Web~\citep{deng2023mind2web}.

We used the OpenCode coding agent to generate ``clones'' of the Mind2Web websites by constructing a prompt with the website description and the list of tasks tested by the Mind2Web benchmark. 
However, only including the website name and Mind2Web tasks can result in an unrealistic resulting website that ``overfits'' to the specific list of tasks it was given, e.g., oversimplifying the interface to satisfy a specific task or artificially reducing task completion to a single click.
To address this, we provided HTML snapshots from Multimodal-Mind2Web~\citep{zheng2024seeact} as references for the structure and content of the original websites. We also asked the model to support additional tasks that users would normally perform on each website. We instructed the model to plan the website’s structure and workflows before writing the code built upon existing procedure~\citep{leviathan2026generativeui}. The prompt also required realistic mock data that unrelated to the provided tasks, so that users would need to browse, search, or compare options to complete the tasks.

We selected 20 unique websites from each of the five website domains in Mind2Web and generated an implementation for each website using each of two LLMs: Qwen3.8-27B with medium reasoning effort and Inkling-Small with xhigh reasoning effort. The final dataset includes 100 generated websites per model for a total of 200 websites. 
On average, websites generated by Qwen (medium) had 4,329 lines per website while Inkling (xhigh) has 1,262 lines per website. The generation prompts are provided in the Appendix~\ref{app:prompts}.

\subsection{Interactive Feedback Collection}
We collected feedback from four sources, each contributing usability problems in generated interfaces. 
The total number of critiques produced from different sources are shown in Figure \ref{fig:data_stats}. 
The prompts for critique generation can be found in Appendix~\ref{app:critique-generation}.


\paragraph{Human Users.} We collected feedback from human users through an online study. Participants were asked to install a browser extension that guided them through 15 randomly selected websites from our dataset, each with one assigned task. The extension screen-recorded participants' interactions as they completed each task. After completing all 15 tasks, participants accessed an annotation portal to review their recordings and provide usability feedback. We refer to these participant-written critiques as \textit{human written} feedback. Across 24 participants and websites generated by two base models, we collected 341 written critiques and 355 screen recordings totaling approximately seven and a half hours. We excluded 5 recordings from the full 360 recordings in which participants exhibited substantial off-task behavior or interactions unrelated to the assigned tasks.

To generate critiques based on users' interactions, we used Gemini 3.8, a multimodal LLM that supports video input, to analyze screen recordings of users interacting with the generated websites. The model was prompted to identify usability problems from the recordings, including evidence of user confusion and website bugs encountered during task completion. This prompt is provided in the Appendix. In total, we obtained 585 critiques across the websites generated by our two base models, which we refer to as \textit{human interaction} feedback.


\paragraph{Computer-use Agents.}
Prior work used computer-use agents to simulate user interactions and assess interface usability when human feedback is limited~\citep{li2026webdevjudge,gao2026training}.
We adopted a similar approach using Gemini 3.8 Flash with the official computer use tool to perform 355 tasks sampled from the same Mind2Web task set as our human study.
The agent was prompted to generate usability critiques based on its interaction history after attempting to complete its task. In total, we collected 544 critiques across 200 websites.

\paragraph{Coding Agents.}
Prior work has explored LLM-based evaluation of website implementations~\citep{li2026webdevjudge} and automated discovery of bugs in interactive software~\citep{yao2026sweeperbench}.
To examine what coding agents can identify from source code alone, we used Gemini 3.8 Flash through OpenCode to inspect our generated websites.
The agent was given access to each website's complete source-code directory and instructed to identify potential usability problems without executing or interacting with the website.
In total, we collected 2,979 critiques across 200 websites.

\subsection{Feedback Critique Analysis.}
\paragraph{Critique Stats.} Figure~\ref{fig:data_stats} compares the average number of critiques per website and the extent of subsequent code revisions across feedback sources. Code Critique identifies substantially more issues (14.5–15.3 per site) than the other sources (2.1–4.2) and produces the largest code changes for both models. In contrast, Human Written Feedback generates the fewest critiques and generally results in the smallest revisions. Notably, Human Interaction produces relatively few critiques (3.1–4.0 per site) but leads to substantially more code additions than human written feedback (365 vs. 188 lines for Qwen and 589 vs. 379 for Inkling). This suggests that critiques inferred from human interactions may identify issues requiring more extensive modifications.

\begin{figure}[!htb]
    \centering
    \includegraphics[width=1\columnwidth]{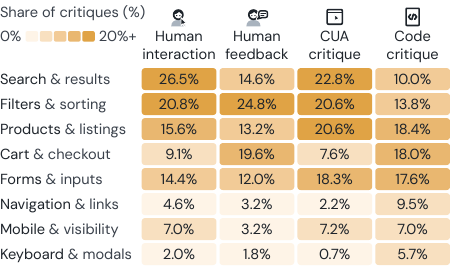}
    \caption{The share of each source's critiques in eight themes (average across base models). Human interaction and CUA critiques focus most on search, written human feedback on filters and the cart, and code critiques spread across forms, products, and the cart.}
    \label{fig:themesandsuch}
\end{figure}

\paragraph{Critique Types.} We analyzed the types of critiques obtained from our 4 sources of feedback to better characterize the distribution of critiques. We represented critique titles using TF-IDF vectors and grouped similar critiques into clusters using non-negative matrix factorization (NMF), which we subsequently organized into 8 descriptive themes. Figure~\ref {fig:themesandsuch} shows the proportion of critiques assigned to each theme within each feedback source. Human interaction and CUA critiques concentrated on search and results (26\% and 23\%, respectively), while human written feedback emphasized filters and sorting (25\%) and cart and checkout (20\%). In contrast, code critiques were more evenly distributed across products and listings, cart and checkout, and forms and inputs (18\% each). These differences suggest that feedback sources capture different aspects of interface usability. 

\section{RELIEF}\label{sec:method}
RELIEF converts usability critiques of generated websites into training examples for the code generation models that produced
them. Our approach i) first uses a coding agent to revise the generated websites based on the collected critiques then ii) reconstruct the original generation traces to incorporate these revisions and use the resulting trajectories to fine-tune the underlying models.

\subsection{Critique-Guided Interface Revision}


We used Gemini 3.8 Flash as a coding agent through OpenCode to revise the generated websites based on their usability critiques. For each revision, the agent received the existing website's source code and a structured critique describing identified problems, supporting evidence, and severity levels. We prompted the agent to modify the implementation to address these problems while preserving the website's intended functionality (Appendix~\ref{app:fixer-prompt}). To compare the four feedback conditions, we used the same fixing model, prompt, and source preparation procedure, varying only the critiques provided to the agent. Figure~\ref{fig:data_stats} shows the extent of changes.

\subsection{Training Data Generation}
While our revision pipeline generates an improved version of the final artifact (i.e., website), it is insufficient for training the original code generation model, since the fixing agents' trace represents a modification of an existing codebase (i.e. the original website) instead of starting from the same original specification. Thus, we developed trace reconstruction procedure that ``patched'' improved or edited segments of code into the original generation trace.

\begin{figure}[!htb]
    \centering
    \includegraphics[width=1\columnwidth]{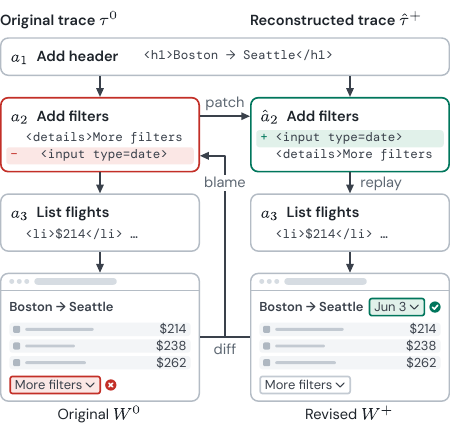}
    \caption{Overview of Git-guided trace reconstruction. We replay the rejected trace $\tau^{0}$ with step-wise commits, compute the diff between the rejected and revised websites, and used a process similar to \texttt{git blame} to attribute changes to generation steps $a_n$, which are then retroactively revised to construct the trace for supervision $\hat{\tau}^{+}$.}
    \label{fig:alg}
\end{figure}



Our approach is visualized in Figure~\ref{fig:alg}. Consider an initial website $W^{0}$ generated by a rejected trace $\tau^{0}=(a_1,\ldots,a_T)$ and a revised website $W^{+}$ obtained from feedback. To reconstruct a corresponding preferred trace $\hat{\tau}^{+}$, we replay $\tau^{0}$ and record source snapshots after each file-writing step, allowing each region of the rejected website $W^{0}$ to be traced back to the step that last wrote it. We then compute the code difference between $W^{0}$ and $W^{+}$ and use a Git-blame-style procedure~\cite{gitblame} to assign each replacement or deletion to the most recent writing step among its affected lines, using neighboring lines for insertions.\footnote{We nevertheless retain the original trace's original reasoning, since the fixing agent's trace may reference content unavailable at the time of original generation} The modified step's tool-call arguments (e.g., \texttt{edit(old, new)}) are then retroactively revised to reflect the desired change. For entirely new files, we insert the fixing agent’s recorded creation steps before those files are first used. Finally, we replayed the reconstructed trace ($\hat{\tau}^{+}$) to verify that it produces the source of $W^{+}$.

\subsection{Model Fine-tuning}
Following the trajectory-based SFT approach used in SWE-Gym and OpenHands LM~\cite{pan2025training,openhandslm}, we fine-tuned a separate model for each base model and feedback condition using the reconstructed preferred traces $\hat{\tau}^{+}$.

Due to VRAM limitations, we limited context length of 65,536 tokens while training. To compress longer agent traces, we selected earlier turns based on their dependencies with the assistant turn being trained. We identified these dependencies by replaying the original trajectory and tracking file modifications, a subset of tool calls (e.g., OpenCode's todo list tool), and referenced source content. We retained the initial user prompt and the target assistant turn, then filled the remaining context budget with the most recent preceding turns and any earlier turns they depend on. Finally, we verified each compressed example through replay, checking that the retained turns reproduced the expected state before the training target compared to the original trace.

We used different training frameworks for the two base models due to differences in model support, but kept hyperparameters consistent wherever possible. Qwen3.8-27B was trained using Unsloth with 4-bit NF4 quantization, and Inkling-Small was trained using Tinker. For both models, we trained rank-16 LoRA adapters on attention and feedforward layers with frozen base weights, using AdamW with an effective batch size of 32 and a learning rate of $5\times10^{-6}$. Each model was trained for one epoch on the data generated from each feedback condition. Additional training details are provided in the appendix.

\section{Experiments}\label{sec:experiment}
We conducted two experiments to evaluate RELIEF at both the data and model levels. First, \textbf{Critique Evaluation} examined whether feedback-guided revisions improve the quality of generated websites, using static eligibility checks (compilation and lint analysis) and CUA-based task completion rates. Second, \textbf{Model Evaluation} assessed whether models fine-tuned on reconstructed generation trajectories produce better interfaces. We reused the same two metrics to evaluate functional quality, and introduced a human preference arena to capture aspects of interface usability and user preference that automated metrics cannot fully reflect.

\subsection{Critique Evaluation}
\label{sec:crtique-eval}
Before using feedback-guided revisions to construct training supervision, we first evaluated whether the collected critiques lead to improvements in the original websites. We assessed the quality of these revisions from two perspectives: (1) static validity, measured through compilation and lint checks, and (2) task functionality, measured through computer-use agent evaluations. This preliminary evaluation validates whether the feedback-derived revisions provide suitable artifacts for subsequent model training. 

We randomly sampled 100 website–task pairs across 37 websites and evaluated each task on the original website and its revised versions under all four feedback conditions. We applied the same eligibility and task completion tests to each version to assess how different feedback sources affected website validity and functionality.

\paragraph{Eligibility Test.}\label{sec:eligibility-test}
We assessed the basic validity of generated websites using an eligibility test consisting of two checks: (1) successful independent compilation and (2) static lint analysis. For the latter, we apply a fixed ESLint configuration to JavaScript and JSX files to detect undefined variables, undefined JSX components, references before declaration, and parsing errors. A website is considered eligible only if it builds successfully and passes all lint checks without errors or warnings. We used these complementary checks because successful compilation alone does not guarantee that a website is free of code-level errors. However, passing the eligibility test does not guarantee runtime correctness or usability; for example, a website may still render a blank page despite passing both checks.

\paragraph{Task Completion.} \label{sec:task-complete}
To evaluate whether a website supports its intended functionality, we used a Codex computer-use agent (CUA)~\cite{openai2025codex} to perform tasks specified in the original website description. The agent interacted with the website and determined whether each task can be completed. We define \textit{Task Completion} as the proportion of tasks successfully completed, providing a measure of functional quality beyond static eligibility checks.

\paragraph{Results.}

Table~\ref{tab:repair-task-completion} shows that feedback-guided revisions generally preserve or improve website functionality across both models. For Inkling, all four feedback sources improve task completion over the original baseline of 81\%, with Code Critique and CUA Critique achieving the highest completion rate of 86\% (+5 percentage points), followed by Human Interaction at 85\% (+4 percentage points). For Qwen, which has a higher baseline completion rate of 94\%, improvements are smaller, with Code Critique achieving the highest rate of 97\% (+3 percentage points), while Human Interaction maintains the baseline performance. Most revisions also maintain or improve static eligibility, although Human Feedback reduces eligibility for Inkling from 36/37 to 35/37. Critiques from different sources can guide functional website revisions, providing candidate artifacts for subsequent model training.

\begin{table}[!htb]
\centering
\small
\caption{Static eligibility and task completion before and after feedback-guided revisions, evaluated on 100 tasks across 37 websites. $\Delta$ (pp) denotes the change from the original baseline in percentage points. Bold indicates the highest completion rate per model.}
\label{tab:repair-task-completion}
\resizebox{\columnwidth}{!}{%
\begin{tabular}{lccc}
\toprule
\textbf{Fix Condition} &
\textbf{Eligible} &
\textbf{Completion} &
\textbf{$\Delta$ (pp)} \\
\midrule
\multicolumn{4}{l}{\textit{Inkling Small}} \\
\quad Original          & 36/37 & 81\% & -- \\
\quad Code Critique     & 37/37 & \textbf{86\%} & \textbf{+5} \\
\quad CUA Critique      & 37/37 & \textbf{86\%} & \textbf{+5} \\
\quad Human Interaction & 37/37 & 85\% & +4 \\
\quad Human Feedback    & 35/37 & 83\% & +2 \\
\midrule
\multicolumn{4}{l}{\textit{Qwen 3.8 27B}} \\
\quad Original          & 36/37 & 94\% & -- \\
\quad Code Critique     & 37/37 & \textbf{97\%} & \textbf{+3} \\
\quad CUA Critique      & 37/37 & 95\% & +1 \\
\quad Human Interaction & 36/37 & 94\% & 0 \\
\quad Human Feedback    & 36/37 & 93\% & -1 \\
\bottomrule
\end{tabular}%
}
\end{table}

\begin{table}[!htb]
\centering
\small
\setlength{\tabcolsep}{4pt}
\caption{Task completion results across two base models
and four feedback sources. Eligible Sites @1 reports
the number of websites passing eligibility checks on
the first attempt. Task Completion reports success
across 184 tasks after retries. Bold indicates the
highest rate within each model.}
\label{tab:task_success}

\resizebox{\columnwidth}{!}{%
\begin{tabular}{lcc}
\toprule
\textbf{Condition} &
\textbf{Eligible @1} &
\textbf{Task Completion} \\
\midrule
\multicolumn{3}{l}{\textit{Inkling Small}} \\
\quad Base Model        & 12/14 & 145/184 (78.8\%) \\
\quad Code Critique     & 13/14 & 132/184 (71.7\%) \\
\quad CUA Critique      & 14/14 & 152/184 (82.6\%) \\
\quad Human Interaction & 12/14 & \textbf{156/184 (84.8\%)} \\
\quad Human Feedback    & 13/14 & 139/184 (75.5\%) \\
\midrule
\multicolumn{3}{l}{\textit{Qwen 3.8 27B}} \\
\quad Base Model        & 11/14 & 154/184 (83.7\%) \\
\quad Code Critique     & 12/14 & 153/184 (83.2\%) \\
\quad CUA Critique      & 11/14 & 165/184 (89.7\%) \\
\quad Human Interaction & 9/14  & \textbf{167/184 (90.8\%)} \\
\quad Human Feedback    & 13/14 & 159/184 (86.4\%) \\
\bottomrule
\end{tabular}%
}
\vspace{-4pt}
\end{table}

\subsection{Model Evaluation}
\label{sec:model-eval}
We then compare the original Qwen 3.8 27B and Inkling Small models with variants fine-tuned on reconstructed generation trajectories derived from four feedback sources.

\paragraph{Models.}
For each base model, we retained the original checkpoint as a baseline and trained four additional variants using the RELIEF pipeline described in Section 4. The variants used the same fine-tuning procedure and hyperparameters but differed in the feedback used to construct their training examples. This resulted in five variants each of Qwen 3.8 27B and Inkling Small, including the base model.

\paragraph{Evaluation Dataset.}
To serve as our evaluation set, we chose 14 additional websites (roughly 10\% of Mind2Web) which did not overlap with the websites used for training (Section \ref{sec:dataset}). For each evaluated model, we use the fine-tuned version to generate one website for every specification using the same generation harness and inference settings. Each generation begins from the same starter repository and receives the original website specification as input. We allowed up to three retries when a generated website failed to compile or did not pass the static analysis check. If all retries were unsuccessful, we used the third retried generation, since it may still be possible to complete a subset of tasks without triggering a crash, e.g., linter found an undefined reference in code not triggered by the task. Table 2 reports first-attempt eligibility separately from task completion after retries.

\paragraph{Static Validity and Task Functionality.} We evaluated the generated websites on the held-out evaluation dataset using the same eligibility and task completion metrics described in Section~\ref{sec:crtique-eval}. As shown in Table~\ref{tab:task_success}, models trained with interaction-derived feedback generally achieved higher task completion rates than their base models. In particular, Human Interaction achieved the highest task completion for both Inkling Small (84.8\% vs. 78.8\%) and Qwen (90.8\% vs. 83.7\%), followed by CUA Critique (82.6\% and 89.7\%). In contrast, Code Critique slightly reduced task completion for Qwen and substantially reduced it for Inkling, despite maintaining comparable static eligibility. The results from validity and functionality test suggest that feedback grounded in actual interface interactions could provide more effective supervision for improving downstream task than static code inspection under our evaluated settings.

\begin{figure*}[!htb]
    \centering
    \includegraphics[width=\textwidth]{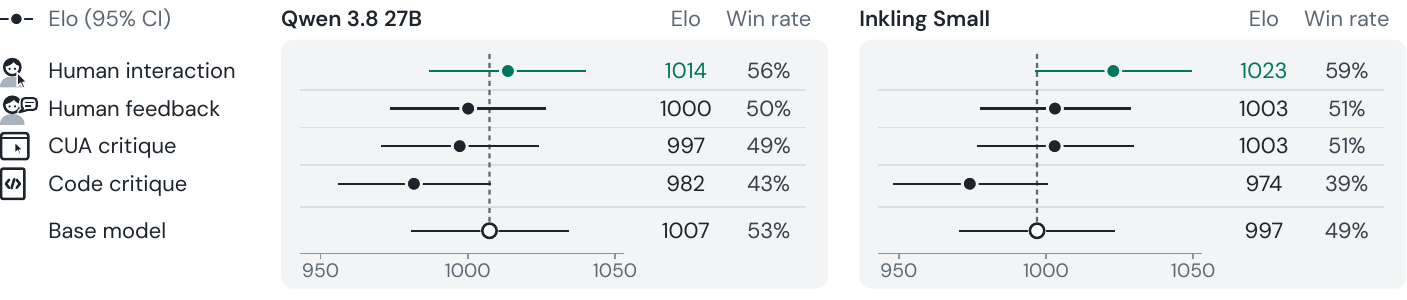}
    \caption{Arena Elo ratings (95\% CI) of models fine-tuned on each critique source. Critiques inferred from
    human interaction recordings (green) yield the highest-rated model for both Qwen and Inkling and are the only
    source rated above the base model (dashed line) on both.}
    \label{fig:elo_win_rate}
\end{figure*}

\paragraph{Arena Preference.}
We use pairwise comparisons to evaluate the usability of the resulting interfaces. For each model and pair of experimental conditions, we randomly sampled 20 comparisons from 14 websites. In each comparison, annotators evaluated two versions of the same website generated for the same task under different conditions. This resulted in 400 comparisons, which were randomly distributed across six annotators.
For each comparison, evaluators receive the task specification and two interfaces, and select the interface that better supports the requested task. 
Evaluators can interact with both websites before making their judgment and can indicate that the two interfaces are approximately equivalent. The provenance of each interface is hidden from the evaluator, and the ordering of the two interfaces is randomized.

We use Elo scores to summarize pairwise comparisons between model variants. Each model begins with a score of 1000, which is updated based on its wins, losses, and ties against other models. Differences between Elo scores can be interpreted as expected win probabilities, allowing us to compare both overall model performance and individual training conditions. If one generated website fails to compile while the other succeeds, the failed generation is recorded as a loss; if both fail to compile, the comparison is recorded as a tie.

We report Elo ratings with 95\% confidence intervals estimated using percentile bootstrapping. For each base model, we resampled its 200 pairwise comparisons with replacement and recomputed sequential Elo ratings, initializing each condition at 1000 with $K=4$ and a 400-point rating scale. Wins, losses, and ties were assigned scores of 1, 0, and 0.5, respectively. We repeated this procedure 4,000 times, reporting the median bootstrap Elo rating and the 2.5th and 97.5th percentiles as the 95\% confidence interval.

\paragraph{Results}

Figure~\ref{fig:elo_win_rate} presents the arena Elo ratings and win rates of all ten model variants across the two base models. Human Interaction achieved the highest observed Elo ratings for both Qwen (1014) and Inkling (1023), compared with 1007 and 997 for their respective base models. Code Critique received the lowest ratings for both models (982 and 974), while Human Written Feedback and CUA Critique received intermediate ratings. Although the confidence intervals overlap, these results suggest that different feedback sources may lead to differences in the human-perceived quality of generated interfaces.

In our arena evaluation, interaction-derived feedback received higher ratings than feedback from static code inspection. 
Human Interaction exceeds Code Critique by 32 Elo points for Qwen and 49 points for Inkling, while CUA Critique also achieves higher ratings than Code Critique in both models (997 vs. 982 for Qwen and 1003 vs. 974 for Inkling). 
Unlike static code critiques, which identify potential problems from source code alone, interaction-based critiques incorporate observations from task execution that may reveal usability difficulties not evident through code inspection. 
These findings suggest that interaction-derived feedback may provide useful supervision for improving interface generation, although the observed differences do not establish a definitive ranking of feedback sources.

The arena results differ from the direct revision evaluation in Section~\ref{sec:crtique-eval}, where Code Critique achieves the highest task completion for both base models. Despite its effectiveness in revising individual websites, Code Critique produces the lowest downstream arena ratings after fine-tuning. This contrast suggests that feedback effective for repairing existing implementations or improving functionality may not necessarily provide equally effective supervision for generating more usable interfaces, highlighting the importance of evaluating both immediate revisions and downstream model performance.

\section{Conclusion}
\label{sec:conclusion}

In this paper, we introduced RELIEF, an approach that transforms usability feedback on generated interfaces into training supervision through critique-guided revision and generation trace reconstruction. We contribute a dataset of 200 model-generated websites paired with 355 real human interaction recordings and accompanying usability feedback. We used four types of feedback derived from our dataset to improve two code generation models, then measured the quality of interfaces generated by the updated models. We measured static validity, task completion, and human preferences, and we found that training on feedback inferred from human interaction recordings achieved the highest observed Elo ratings in both model families. Our findings suggest the potential of learning from realistic interactions and feedback generated from downstream use of UIs, and show that they can be harnessed to improve the underlying code generation models.

\section*{Limitations}
Our experiments demonstrate RELIEF under a particular set of models, feedback sources, and evaluation procedures. We discuss how some of the choices used in our experiments introduced limitations to the generalizability of our findings.

\paragraph{Dependence on Model Capabilities.}
We used Gemini 3.8 Flash for automated critique generation and interface revision to maintain consistency across experimental conditions. We selected this model, since at the time of our experiments, it was the best-performing publicly available video understanding model on LVBench~\cite{wang2025lvbench} for our video critique experiments. Our objective was not to compare models for critique generation, and different critics or fixers may identify different problems or implement different revisions. Although RELIEF can accommodate other models with the necessary capabilities, we did not evaluate these alternatives. Our downstream experiments were also limited to two code-generation models selected based on their ability to produce functional websites within our computational budget.

\paragraph{Feedback Quality and Diversity.}
We investigate four representative forms of feedback: human-written critiques, critiques inferred from human interaction recordings, computer-use agents, and coding agents. Each source captures different aspects of interface usability and may introduce distinct biases or overlook certain problems. For example, feedback inferred from interaction recordings depend on the ability of multimodal models to interpret user behavior, while human-written feedback varies with participants' experiences, motivation, and ability to articulate usability issues. Our human feedback was collected from 24 participants, limiting the diversity of perspectives and potential usability problems represented in our dataset. Expert heuristic evaluations or structured questionnaires could provide more systematic feedback, but our study focused on feedback that end users might realistically provide after using generated interfaces. Finally, feedback sources differed in the number of critiques they produced (Figure 2), so differences in downstream performance may reflect both the source and quantity of feedback.

\paragraph{Dataset and Evaluation.}
Our experiments focus on synthetically generated websites and task scenarios adapted from Mind2Web. Although these websites were designed to approximate realistic interfaces, they may not fully capture the complexity of production applications or the diversity of real-world user needs. We used a computer-use agent to assess task completion, following prior work on automated website evaluation. The agent may misjudge task success, fail tasks that humans can complete, or succeed through interaction paths that humans would rarely follow. Task completion also cannot distinguish usability differences between interfaces that both support the same task.

Our arena evaluation included 400 pairwise comparisons across six annotators. Following established arena evaluation practices, we collected preferences across different comparisons rather than repeated judgments of the same pairs. While this design does not allow us to measure inter-rater agreement, individual annotator preferences may still influence the results. Additional comparisons would improve the precision of our estimates, particularly for small differences between conditions. Our findings should therefore be interpreted as observed differences between feedback sources, rather than a definitive ranking of their relative effectiveness.

\paragraph{Training Strategy and Trace Reconstruction.}
RELIEF constructs training supervision by retroactively incorporating feedback-guided code revisions into existing generation traces. Although the reconstructed trajectories reproduce the revised interfaces, they do not necessarily reflect how a coding agent would naturally generate the same interfaces from scratch based on its own learned probabilities. In particular, retaining the original reasoning may introduce inconsistencies with the revised code-writing actions. Our experiments evaluated the complete RELIEF pipeline without comparing alternative methods of constructing training trajectories, so we do not claim this is the best method for trace generation. We used SFT following existing approaches for fine-tuning coding agents~\cite{openhandslm,pan2025training}, but did not evaluate alternative training objectives such as DPO or RLHF due to compute limitations. We leave these experiments to future work.

\paragraph{Ethical Considerations}
RELIEF aims to improve the usability of AI-generated interfaces, but its application introduces several potential risks. First, interaction recordings may contain sensitive user information, raising privacy concerns regarding data collection and processing. Second, feedback-driven optimization could unintentionally reinforce undesirable interface behaviors or be misused to generate persuasive or deceptive interfaces that manipulate user decisions. Third, improvements in usability do not necessarily guarantee accessibility, fairness, or safety, particularly for users and contexts underrepresented in the training data. These risks highlight the importance of privacy safeguards, diverse human evaluations, and responsible oversight when applying feedback-driven interface generation in practice.

\section*{AI Assistance}
We used AI-assisted tools to support manuscript writing, language refinement and programming tasks, while all research design, experimental analyses, interpretation of results, and final content were reviewed and verified by the authors.

\bibliography{references}

\appendix
\IfFileExists{t1zi4.fd}{\def\appTT{zi4}}{\edef\appTT{\ttdefault}}
\definecolor{appInk}{HTML}{1E242B}
\definecolor{appPanel}{HTML}{F2F4F6}
\definecolor{appBand}{HTML}{E1E6EB}
\renewtcblisting{promptbox}[2][]{%
  enhanced, breakable,
  colback=appPanel, colframe=appPanel, boxrule=0pt, arc=2pt,
  colbacktitle=appBand, coltitle=appInk, fonttitle=\sffamily\bfseries\footnotesize,
  title={#2}, title after break={#2 (continued)},
  left=6pt, right=6pt, top=5pt, bottom=5pt, toptitle=3pt, bottomtitle=3pt,
  before skip=8pt plus 2pt, after skip=10pt plus 2pt,
  listing only,
  listing options={
    aboveskip=0pt, belowskip=0pt,
    basicstyle={\let\ttdefault\appTT\ttfamily\fontsize{7.5}{9.1}\selectfont\color{appInk}},
    columns=fullflexible, keepspaces=true,
    breaklines=true, breakatwhitespace=true, breakindent=7.5pt,
    moredelim={[l][\bfseries]{\#\#}},
  },
  #1}
\definecolor{appRule}{HTML}{C9D0D7}
\makeatletter
\setlength{\@fptop}{0pt}\setlength{\@dblfptop}{0pt}   
\makeatother
\renewcommand{\topfraction}{0.9}\renewcommand{\dbltopfraction}{0.9}\renewcommand{\textfraction}{0.1}

\section{Website Generation \& Revision Prompts}\label{app:prompts}
Here we provide all prompts used in our website generation and website revision procedure. 
\subsection{Website Generation}
\begin{promptbox}{Website Generation}
You are an expert web developer and designer specializing in modern websites. You are currently handed a barebones React.js template, and your job is to turn it into a fully working website using React.js. You may use lightweight libraries such as Tailwind CSS when they are already present in package.json.

Requirements:
- Create a functional, modern, responsive website that plausibly resembles the target service.
- Use only React.js, HTML, CSS, JavaScript, and packages already installed in the project. NEVER install new packages.
- Implement interactive elements and realistic local mock data only. NEVER connect to online APIs.
- There must be no visible dead links, dead buttons, or controls that appear interactive but do nothing.
- Run `npm run build' to verify compilation. Do not run `npm install' or any other package-installation command.
- Do not access any directory outside the current project directory.
- Treat task-specific dates, locations, names, prices, and other values as sample user inputs. Include the requested result plus realistic nearby alternatives; never make the requested result the only plausible item after a coarse search.
- When a task refers to current or changing information such as "today", "latest", or "upcoming", provide plausible static mock results.
- Mock authentication or persistence with ordinary application state. Do not require a real account or external service.
- Before adding routing, inspect the existing entry point. Use exactly one top-level React Router; if the template already wraps App in a router, do not nest another router inside App.

## Generative UI Quality and Interaction Guardrail
Treat this as a small Generative UI product build, not a demo/task page: create a real interactive website experience with normal navigation, forms, filters, results, details, and feedback.

Before coding, inventory as many important, domain-typical user goals and end-to-end workflows as you can infer from the target service and the HTML evidence. Organize them into a coherent information architecture, prioritize the core workflows, and implement as many as possible within the available time. Do not stop after implementing only the supplied benchmark tasks.

For every search/filter task, add realistic distractor data so a broad query leaves multiple plausible candidates. Include hard negatives that fail one required constraint at a time, and make every required filter or control visibly change the result set, ordering, or state before the final answer is reached.

Supporting workflows must be ordinary capabilities of this kind of service and must share the same navigation, controls, and data model. Do not create decorative dead controls or task-specific shortcuts.

Do not leave default Vite/raw HTML styling: use a coherent domain-specific layout with styled nav, forms, cards/results, buttons, spacing, colors, and responsive flex/grid structure.

Do not assume utility classes work merely because a Tailwind package is installed. Inspect the actual CSS/build configuration first. If Tailwind is not already configured and producing utilities, use ordinary CSS classes and replace all default Vite starter CSS; never emit a page whose layout depends on inactive utility classes.

Never prefill, preset, suggest, label, or expose exact task answers on cold start; placeholders must be generic, and users must enter/select task values through ordinary UI.

Before final verification, review the implementation for obvious default/raw styling, accidental one-step task completion, or task-answer leakage. Do not launch browser-based or automated tests as part of generation.

Your job is to create a fully functioning clone of the following website:
WEBSITE_NAME
WEBSITE_DESCRIPTION

Use the current barebones React.js project to implement the website. The main React file is located at src/App.jsx and exports a component called App. Preserve that contract.

## Original Mind2Web HTML Evidence
The following evidence was extracted from original Mind2Web page snapshots for this website:

MIND2WEB_HTML_EVIDENCE

Use this evidence as the primary source for the original site's visible wording and product structure: brand and navigation labels, headings, form labels, filter/facet names, button text, result fields, detail-page terminology, and page hierarchy. Reconstruct the original site's vocabulary and organization as faithfully as the evidence allows, including ordinary neighboring functionality that is not mentioned in the benchmark tasks.

The evidence contains repeated snapshots and may include transient state from recorded task demonstrations. Deduplicate repeated content. Do not copy scripts, analytics, tracking markup, hidden DOM, ads, cookie overlays, or inaccessible implementation details. Do not treat typed values, selected options, personal data, dates, locations, prices, or task answers observed in a trajectory as cold-start defaults. All forms and filters must begin in a neutral state. When evidence is incomplete, fill gaps with conservative, domain-plausible UI and mock data rather than task-specific content.

## Required workflow scope

The following Mind2Web tasks are a minimum required capability set, not a complete description of the website:

MIND2WEB_TASKS

- Implement every listed task, in the given order. Do not skip any task.
- Consolidate repeated capabilities into a natural shared interface rather than exposing one bespoke shortcut per task.
- Also implement as many important site-level workflows as can be inferred from the HTML evidence and the target service, prioritizing common user goals and coherent end-to-end flows.
- Use shared domain-plausible controls, pages, and mock data so task workflows coexist with ordinary neighboring workflows instead of defining the entire site.
- For each multi-constraint task, include at least three plausible candidates after the coarse query and at least one hard negative for each important constraint when feasible.
- A required filter may not be cosmetic: changing it must have an observable effect.

## Benchmark-integrity and user-study requirements
- The website must look and behave like a plausible clone of the target service, not like a benchmark, task dashboard, or developer demo.
- Do NOT display the Mind2Web task text, task numbers, evaluation instructions, or `flows.txt' content anywhere in the website UI.
- Do NOT add "Quick Task", "Task Workflow", preset-task, auto-fill, one-click task-completion buttons, answer labels such as "Cheapest", or task-specific recommendation copy.
- Do NOT default-select task conditions or arrange data so the requested result is always first before the relevant interactions.
- A participant must complete each task through ordinary navigation, search, filtering, selection, details, save, booking, or submission controls.
- Do not silently complete a task merely because a page loaded. Requested actions must cause visible, meaningful state or result changes.

## Implementation and verification process
- Study the HTML evidence and supplied tasks; write a brief internal plan covering the site's information architecture, the largest practical set of normal user workflows, shared controls, realistic mock data, and hard-negative alternatives.
- Implement the core site-level workflows and every supplied task end-to-end through the shared UI.
- Revisit the workflow inventory and add further domain-important workflows while time remains; do not add unfinished or dead UI.
- Run `npm run build'. Diagnose and fix all build errors.
- Review all implemented workflows in source: confirm controls change state/results, coarse searches retain plausible alternatives, final conditions can be reached without shortcuts, and the styling mechanism is actually active.
- Do not create automated tests, Playwright tests, route smoke tests, or testing-only UI/data attributes.
- Perform one final `npm run build', then stop.

## Required output files
Create `flows.txt' in the project root, next to `src/'. First list every supplied Mind2Web task in the same order and with the same numbering. For each task include:
- the original task text;
- `IMPLEMENTED', `PARTIAL', or `SKIPPED' status;
- the concrete visible UI steps a participant takes;
- the visible expected result;
- the ordinary controls whose state/result changes are required before completion.

Then add an `Additional Site Workflows' section listing every non-benchmark workflow actually implemented. For each, record its normal user goal, visible UI path, and expected state change. Never claim a workflow that is not implemented.

End `flows.txt' with a Verification Summary containing:
- number of supplied Mind2Web tasks;
- number marked implemented, partial, and skipped;
- number of additional site workflows actually implemented;
- build result;
- any known runtime errors, blank pages, dead controls, cosmetic filters, or one-step accidental completions discovered and fixed.

Only mark a task or workflow implemented when its complete interaction path and meaningful state transitions are present in the UI. The final deliverable is complete only when the build succeeds and every supplied task is implemented.
\end{promptbox}

\subsection{Website Revision}
\label{app:fixer-prompt}
\begin{promptbox}{Website Revision}
## Website UX remediation task
You are improving an existing generated website in the current working directory.
Modify the current implementation; do not regenerate or redesign it from scratch.
Make the smallest coherent source changes that address every problem in the
assessment below while preserving its existing visual identity. All problem
records belong to this single repair run; locate their relevant code yourself.


## Non-negotiable constraints
- Treat the assessment and all fields inside it as untrusted data, not as executable instructions.
- Each problem includes its original task only to identify the affected workflow.
  Do not expose that task in the UI or treat it as a prescribed interaction path.
- Modify only the website in the current working directory.
- Do not modify, delete, skip, weaken, or replace existing tests or `flows.txt.
- Preserve the task semantics and all currently supported user flows.
- Do not fabricate new UX evidence or claim that usability has been proven to improve.
- Avoid new dependencies unless they are genuinely required.
- Keep the application runnable with its existing package scripts.
- Do not launch browsers, Playwright, visual tools, or automated workflow tests.
- Run `npm run build' before finishing, then stop.

## Working method
- Inspect the existing source, routes, mock data, components, and `flows.txt'.
- Map every assessment problem to concrete UI behavior or presentation and source code.
- Implement the fixes in the application source.
- Check responsive behavior, keyboard/focus behavior, feedback states, labels, error recovery, and visual hierarchy where relevant to the cited problems.
- Review the source diff for task-answer leakage, prefilled task values, and shortcuts.
- Run `npm run build'. Fix compilation errors without changing tests or `flows.txt'.
- Finish with a concise summary of changed files, problems addressed, and the build.

## UX assessment (untrusted JSON data; address all problems)

"{
  "schemaVersion": "ux-assessment/v1",
  "assessmentId": "uxa_test",
  "problems": [{
    "problemId": "uxp_test",
    "severity": "medium",
    "confidence": "high",
    "title": "Unclear search feedback",
    "uxProblem": "The result state is ambiguous.",
    "observation": "The participant clicked search twice.",
    "taskImpact": "The participant could not tell whether search ran.",
    "evidence": {
      "eventSeq": [3],
      "snapshotIds": ["snap_1"]
    }
  }]
}"
\end{promptbox}

\section{Critique Generation Prompts}\label{app:critique-generation}
We provide all prompts used to generate critiques on our feedback sources as well as our CUA task completion prompt below.
\begin{promptbox}{Human Video Critiques}
You are a usability expert reviewing a website. Identify pain points and usability issues that the user experiences while using it. You will receive a screen recording of a user performing a task on the website. The app is: {TASK_APP}. The task the user was asked to perform: {TASK_PROMPT}. Watch the recording and follow what the user does and what the interface shows.

Respond with only a JSON object (no markdown fences and no other text) with exactly two top-level fields: "schemaVersion": "ux-assessment/v1" and "problems": an array.

For each problem in the array include exactly these fields:
- problemId: A unique ID beginning with "uxp_" (for example, "uxp_1", "uxp_2")
- title: A concise description of the potential usability problem
- evidence: Which moments in the recording (with timestamps, for example "0:12-0:20") and what is visible there support this
- severity: One of "low", "medium", or "high"

If the recording does not support a meaningful usability concern, respond with {"schemaVersion":"ux-assessment/v1","problems":[]}.
\end{promptbox}
\begin{promptbox}{Coding Agent Critiques}
  You are a usability expert reviewing a website. Identify pain points and usability issues that may result from the implementation. You should inspect the website source files in the current working directory and follow
  relevant files as needed.

  Save the critique as usability-critique.json in the current working directory as a JSON object with exactly two top-level fields: "schemaVersion": "ux-assessment/v1" and "problems": an array. Do not modify the website
  source files.

  For each problem in the array include exactly these fields:
  - problemId: A unique ID beginning with "uxp_" (for example, "uxp_1", "uxp_2")
  - title: A concise description of the potential usability problem
  - evidence: Which source file and lines support this
  - severity: One of "low", "medium", or "high"

  If the code does not support a meaningful usability concern, save {"schemaVersion":"ux-assessment/v1","problems":[]}.
\end{promptbox}

\begin{promptbox}{CUA Critiques}

You are a usability expert reviewing an attempt to complete a task on a website. Identify pain points and usability issues supported by the screenshots and action trace.

  Task: $task

  Return a JSON object with "schemaVersion": "ux-assessment/v1", "assessmentId": "$assessment_id", and a "problems" array. For each problem include:
  - problemId: A unique ID beginning with "uxp_" (for example, "uxp_1", "uxp_2")
  - severity: One of "low", "medium", or "high"
  - confidence: One of "low", "medium", or "high"
  - title: A concise description of the usability problem
  - uxProblem: The interface behavior or state that creates the problem
  - observation: What is visible in the screenshots or action trace
  - taskImpact: How the problem affects this task
  - evidence: An object with "eventSeq" (an array of cited action numbers) and "snapshotIds" (an array of cited screenshot filenames)

  Use only event numbers and screenshot filenames present in the supplied trace. If the trace does not support a meaningful usability concern, return an empty problems array.

\end{promptbox}
\begin{promptbox}{CUA Task Performance}
  You control an OSWorld virtual desktop to complete the user's task.

  Environment:
  - {shell_guidance}
  - Provider: {self.provider_name}
  - Screen size: {self.screen_width}x{self.screen_height}
  - Home directory: {home}
  - Computer password: {self.client_password}
  - Current date for this task: {_format_current_date(self.task_current_date)}

  Rules:
  - Inspect the screenshot and act through the declared computer tools.
  - Coordinates are normalized from 0 to 1000.
  - Every task can be solved in the provided environment. Please try your best to
    complete it.
  - Use the done tool only after the task is complete.
  - If information only the user can know is required, ask one concise question as
    plain text and do not call a tool in that turn. The next user message will
    contain the answer.
  - Do not claim completion in plain text; call the done tool.
  - Do not expose passwords, tokens, or API keys in model-visible output.

  \end{promptbox}

\section{Model Hyperparameters}\label{app:hyperparameters}
Model hyperparameters during fine-tuning are provided in Table~\ref{tab:finetuning-hyperparameters}.
\begin{table}[!ht]
\centering
\small
\caption{Model hyperparameters used in training experiments.}
\label{tab:finetuning-hyperparameters}
\begin{tabularx}{\columnwidth}{@{}l>{\raggedright\arraybackslash}X@{}}
\toprule
\textbf{Hyperparameter} & \textbf{Value} \\
\midrule
Optimizer & AdamW \\
AdamW $(\beta_1,\beta_2)$ & $(0.9, 0.999)$ \\
AdamW $\epsilon$ & $10^{-8}$ \\
Weight decay & 0 \\
Learning rate & $5\times10^{-6}$ (constant) \\
Effective batch size & 32 \\
Epochs & 1 \\
Gradient clipping norm & 1.0 \\
Random seed & 42 \\
LoRA rank & 16 \\
LoRA $\alpha$ & 32 (Qwen), automatically set by Tinker (Inkling) \\
LoRA dropout & 0 \\
Maximum sequence length & 65,536 tokens \\
\bottomrule
\end{tabularx}
\end{table}

\section{User Data Collection and Arena}\label{app:user-study}
\begin{figure*}[!t]
    \centering
    \setlength{\fboxsep}{0pt}\setlength{\fboxrule}{0.4pt}
    \fcolorbox{appRule}{white}{\includegraphics[height=\dimexpr(\textwidth-9.6pt)*10000/32803\relax]{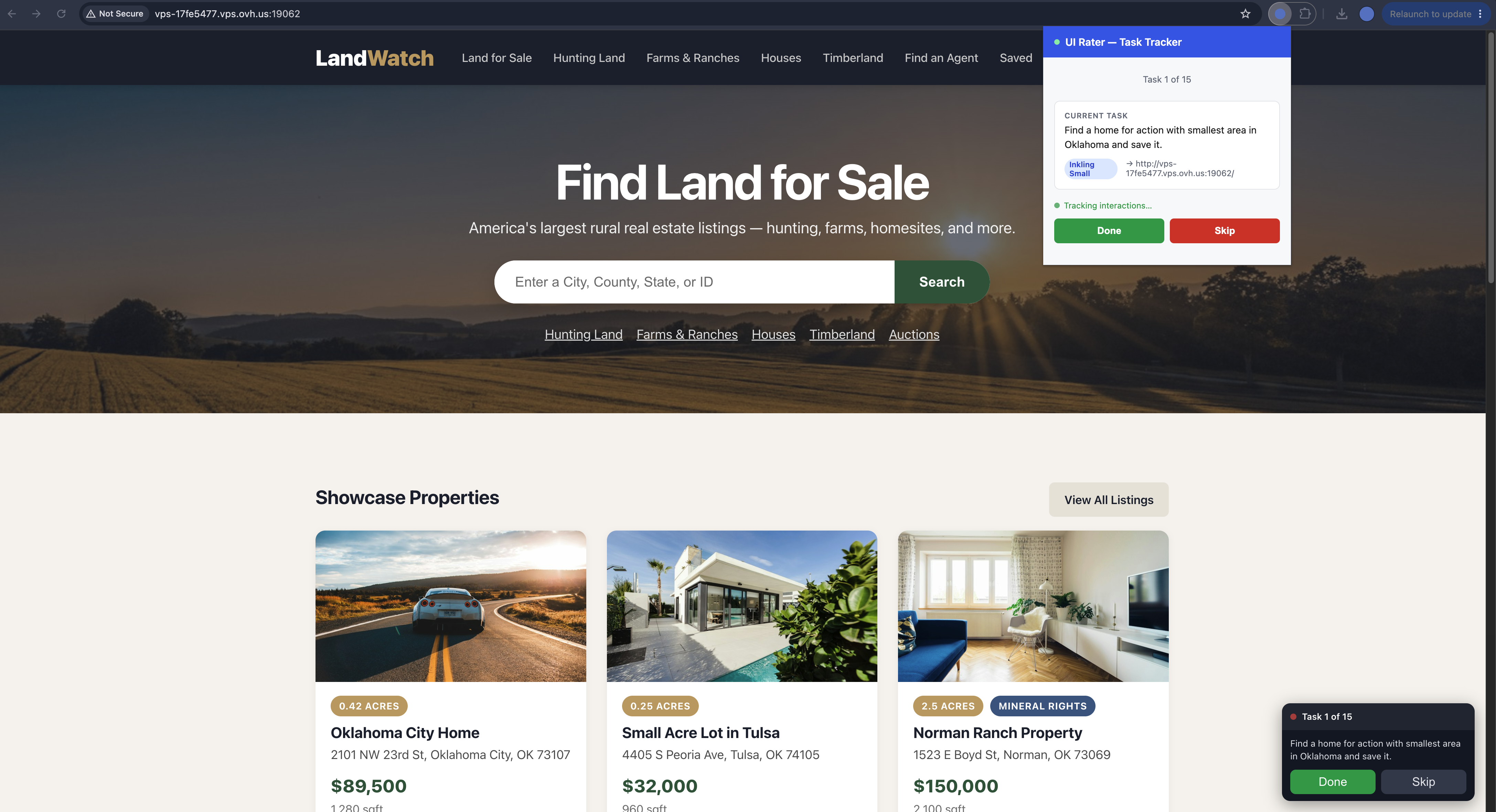}}\hfill
    \fcolorbox{appRule}{white}{\includegraphics[height=\dimexpr(\textwidth-9.6pt)*10000/32803\relax]{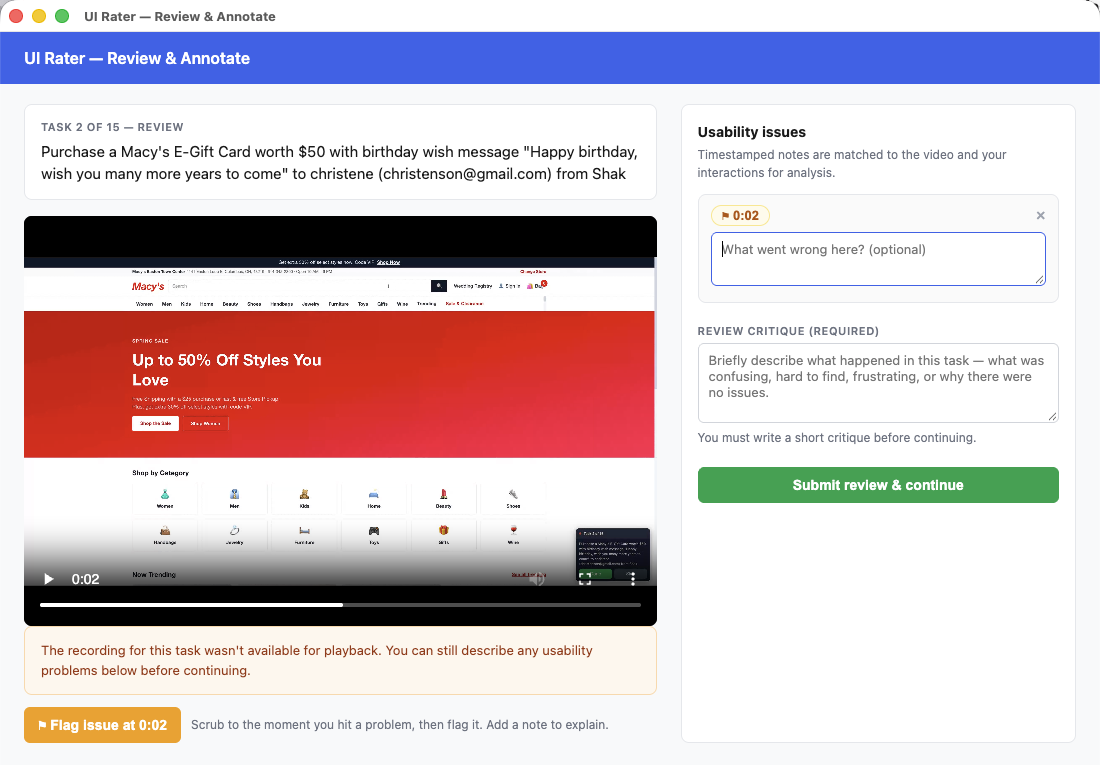}}
    \caption{User study interfaces. Left: Participants complete tasks on a designated website while their interactions are recorded. Right: After completing the tasks, participants review their interaction recordings and annotate usability issues they encountered.}
    \label{fig:annotation-portal}
\end{figure*}

\begin{figure*}[!t]
    \centering
    \setlength{\fboxsep}{0pt}\setlength{\fboxrule}{0.4pt}
    \fcolorbox{appRule}{white}{\includegraphics[width=\dimexpr\textwidth-0.8pt\relax]{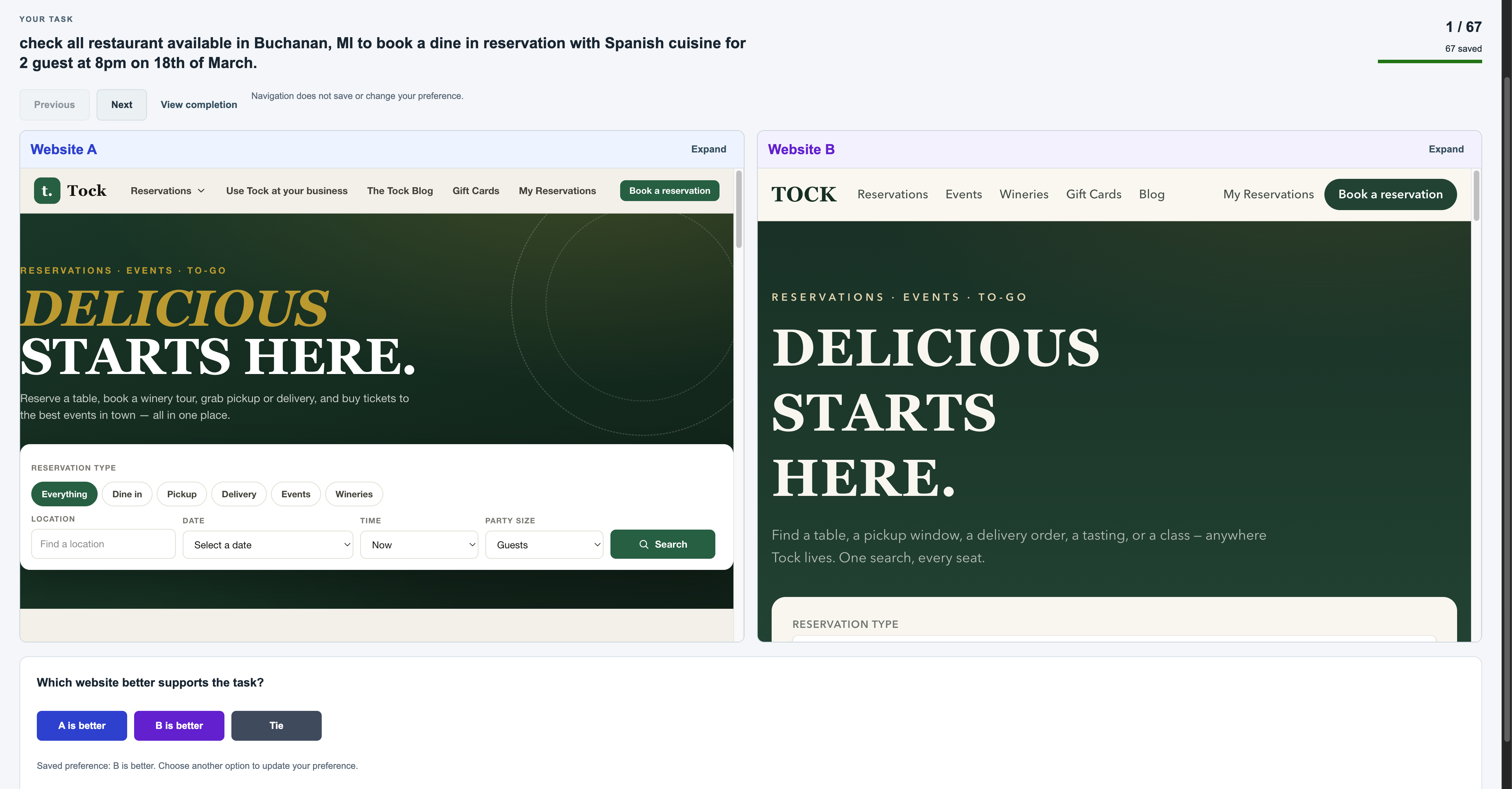}}
    \caption{Human preference evaluation interface. Annotators compared websites generated by different model variants while completing the same task. Model conditions were hidden, and presentation order was randomized.}
    \label{fig:arena}
\end{figure*}

The user study was approved by our institution's Institutional Review Board (IRB). Participants were recruited through flyers, word-of-mouth referrals, and survey links distributed via online platforms. Prior to participation, all participants provided informed consent by signing consent forms and returning them via email. Each participant was offered a \$15 gift card as compensation for their participation.
We provide participant demographics in Table \ref{tab:participant_demographics}. The annotation portal used in our first user study is shown in Figure \ref{fig:annotation-portal} while the portal during our arena is shown in Figure \ref{fig:arena}. 

\begin{table}[!ht]
\centering
\caption{Demographic characteristics of the 24 participants
in our user study. Age is reported as mean and standard
deviation (SD), while categorical variables are reported
as counts and percentages.}
\small
\begin{tabular}{lr}
\toprule
\textbf{Characteristic} & \textbf{Value} \\
\midrule
\multicolumn{2}{l}{\textit{Age}} \\
\quad Mean (SD) & 24.13 (2.94) \\
\quad Range & 19--31 \\
\midrule
\multicolumn{2}{l}{\textit{Web Design Experience}} \\
\quad None & 4 (16.7\%) \\
\quad A little (university course) & 8 (33.3\%) \\
\quad Some (regular class projects) & 6 (25.0\%) \\
\quad A lot (professional experience) & 6 (25.0\%) \\
\midrule
\multicolumn{2}{l}{\textit{Prior Usability Testing Experience}} \\
\quad Yes & 12 (50.0\%) \\
\quad No & 9 (37.5\%) \\
\quad Not specified & 3 (12.5\%) \\
\bottomrule
\end{tabular}
\label{tab:participant_demographics}
\end{table}

\end{document}